# Combining SLAM with muti-spectral photometric stereo for real-time dense 3D reconstruction

Yuanhong Xu, Pei Dong, Junyu Dong*, Lin Qi*

**Abstract**— Obtaining dense 3D reconstruction with low computational cost is one of the important goals in the field of SLAM. In this paper we propose a dense 3D reconstruction framework from monocular multispectral video sequences using jointly semi-dense SLAM and Multispectral Photometric Stereo approaches. Starting from multispectral video, SALM (a) reconstructs a semi-dense 3D shape that will be densified;(b) recovers relative sparse depth map that is then fed as prioris into optimization-based multispectral photometric stereo for a more accurate dense surface normal recovery;(c)obtains camera pose that is subsequently used for conversion of view in the process of fusion where we combine the relative sparse point cloud with the dense surface normal using the automated cross-scale fusion method proposed in this paper to get a dense point cloud with subtle texture information. Experiments show that our method can effectively obtain denser 3D reconstructions.

**Index Terms**—SLAM, 3D reconstruction，Mutispectral Photometric Stereo, Cross-scale fusion

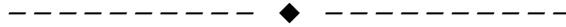

## 1 INTRODUCTION

Real-time 3D reconstruction methods have become increasingly popular research topics among which monocular Simultaneous Localization and Mapping (SLAM) methods are the most advanced for their use in robotics, in particular to navigate unmanned aerial vehicles (UAVs) [1, 2, 4] and augmented and virtual reality applications.

In monocular SLAM where scene geometry is retrieved from a series of images of different views, the representation of geometrical uncertainties is essential. However, uncertainty propagation quickly becomes intractable for large degrees of freedom. This difficulty has split mainstream SLAM approaches into three categories: sparse, semi-dense and dense SLAM methods. Sparse SLAM methods [5, 10, 11] track a set of image feature points to solve camera motion and build a sparse 3D map from these tracked points thus it can't help understand the world more specifically and accurately. Dense SLAM methods [12] produce dense depthmaps and it is important in many applications such as object detection, recognition, obstacle avoidance. But it is computationally demanding and requires a state-of-the-art GPU to run in real-time. Semi-dense SLAM methods which only use high gradient image pixels, in particular along edges, to make tracking more robust and produce relative dense geometry [14, 17, 18] allow real-time operation on a CPU and even on a modern smartphone [19]

The robustness as well as grater utility of dense SLAM and the efficiency of semi-dense SLAM together motivate us to find solutions that have high robustness and low computation cost for denser 3D recovery. A feasible optimization strategy is to densify the semi-dense 3D reconstruction results obtained in real-time SLAM systems using other information. Similar optimization methods have been used in non-real-time systems where they combine different algorithoms and information for dense and accurate reconstruction. [15] uses photometric stereo (PS) to estimate normal and then uses the normal to refine 3D model calculated from a multi-illumination muti-view stereo (MVS) algorithm which is for accurate 3D shape. [13] merges the 3D sparse model and the dense 3D surface obtained with structure from motion (SfM) and PS respectively to reduce the bas-relief ambiguity effect for accurate 3D reconstruction. However, these methods are difficult to directly extend to real-time systems because they can't execute real-time data acquisition or algorithm running.

The essence of above methods is to use additional parameters obtained from other methods to constrain or complement the baseline method. In this paper, we propose to combine real-time semi-dense SLAM and multispectral photometric stereo to optimize the semi-dense reconstruction results to obtain denser 3D recovery. To the best of our knowledge, we are the first to propose the method of combining multispectral photometric stereo and SLAM for real-time 3D reconstruction. Specifically, we insert surface normal calculated from multispectral photometric stereo as additional optimization factors into

---

- Yuanhong Xu. is with the Ocean University of China 266100, Qingdao, P. R. China. E-mail: xuyuanhong@stu.ouc.edu.cn
- Pei Dong is with the Ocean University of China 266100, Qingdao, P. R. China. E-mail: dongpei@stu.ouc.edu.cn.
- Junyu Dong is with Ocean University of China 266100, Qingdao, P. R. China. E-mail: dongjunyu@ouc.edu.cn.
- Lin Qi is with Ocean University of China 266100, Qingdao, P. R. China. E-mail: qilin@ouc.edu.cn.

*Correspoinding Author

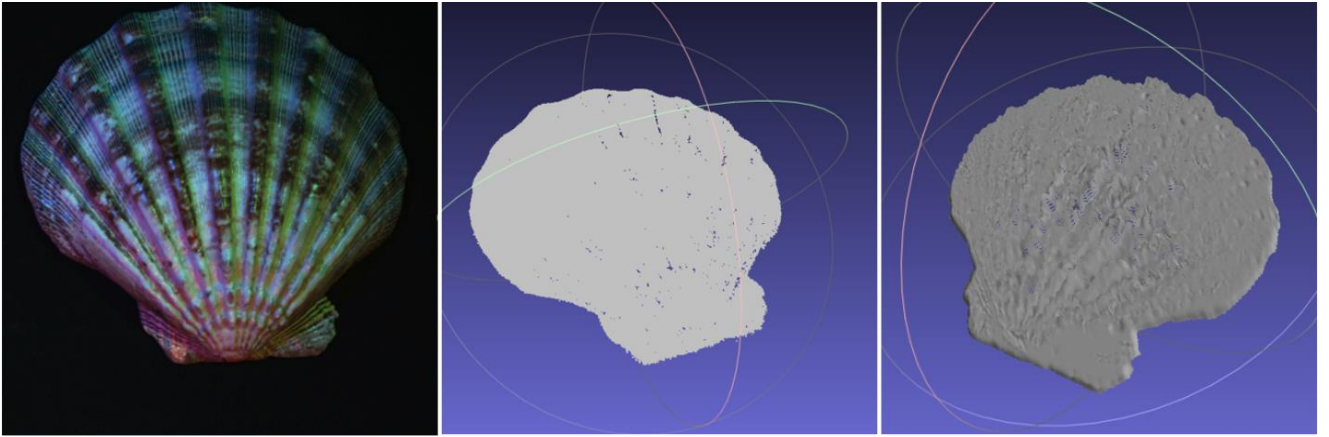

Fig. 1. The results of our framework. Semi-dense SLAM builds relative sparse 3D reconstruction. We insert dense surface normal calculated using multispectral photometric stereo algorithm into semi-dense point cloud to perform joint optimization and we can derive dense 3D model. left: multispectral image; medium: semi-dense point cloud obtained from SLAM; right: dense point cloud using our framework.

semi-dense point cloud obtained from SLAM to perform optimization. Our experiments are performed on video sequences of Lambert model objects, filmed under spatially separated red, green, and blue light sources. The semi-dense SLAM performs tracking, depth estimation and map optimization based on the acquired multispectral data and outputs a relative sparse point cloud, which, as a basic 3D model will be densifed in the next procedure. Apart from the relative sparse point cloud, we can also obtain rotation matrix and translation vector that jointly denotes the camera motion. The camera motion metrix is used to convert point cloud view. Also, we can derive depth maps of each keyframe from semi-dense SLAM which will be used as priori information for optimization-based multi-spectral photometric stereo algorithm [9] to recover a dense surface normal. Then, the dense surface normal from multispectral photometric stereo is added into the semi-dense point cloud from SLAM for further fusion. Afterwards, through area matching and joint optimization, we can get dense point cloud with sutle textures. In addition, we explored the application of ICP algorithms in optimizing fusion results. It is worth mentioning that multispectral photometric stereo can recover pixel-wise surface normal from a single multispectral image which ensures that the rebuild process can be done on the video. And we only perform normal reconstruction of multispectral images selected by SLAM as keyframes, in which way we can save computational costs and increases efficiency. By applying above methods, our framework can conduct a dense reconstruction (see Fig.1). Last but not least, this framework can run in real-time on CPUs since the two processes of semi-dense building using SLAM and dense surface normal recovery using multispectral photometric stereo are simultaneously carried out on two parallel systems.

We performed experiments on different objects. The experimental results show that our proposed framework can obtain dense reconstruction results from semi-dense point clouds with detailed textures. Our main contributions are:

a. We first propose to combine multispectral photometric stereo with SLAM;
b. We propose a novel real-time dense three-dimensional reconstruction method.

## 2 RELATED WORK

In this Section we review related work with respect to the three fields that we integrate within our framework, i.e. SLAM, Multispectral Photometric Stereo and Fusion Stragetry.

### 2.1 SLAM

There exists a vast literature on SLAM. According to the occasion of input data being processed, approaches can be classified into sparse, dense, semi-dense SLAM methods.

Sparse SLAM algorithms usually split the overall problem–estimating geometric information from images – into two sequential steps: First, a set of feature observations is extracted from the image. Second, the camera position and scene geometry are computed as a function of these feature observations only. While this decoupling simplifies the overall problem, it comes with an important limitation: Only information that conforms to the feature type can be used. In particular, when using keypoints, information contained in straight or curved edges – which especially in man-made environments make up a large part of the image – is discarded. Several approaches have been made in the past to remedy this by including edge-based [20,21] or even region-based [22] features. In terms of their application, sparse representations capture only partial scene information and are mainly used for localisation only.

Recently, dense SLAM that makes use of all of the data in an image is increasingly popular because it is possible to get more complete, accurate and robust results. Dense methds exploit all the information in the image, even from areas where gradients are small; thus, they can outperform feature-based methods in scenes with poor tex-ture, defocus, and motion blur. Dense maps capture complete surface shape and can be augmented with semantic labels, but their high dimensionality makes them computationally costly to store and process, and unsuitable for rigorous probabilistic inference.

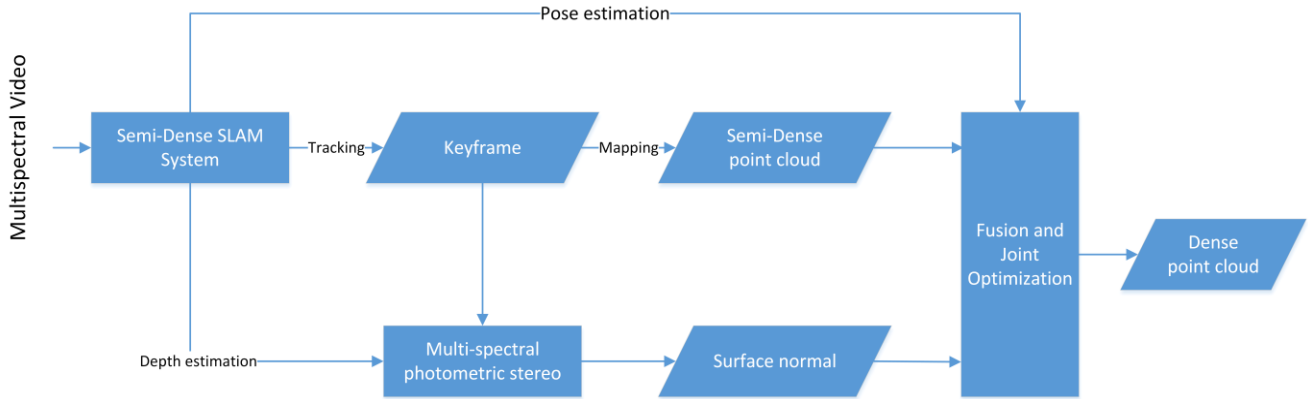

Fig. 2. Overview of the proposed framework.

Semi-dense methods overcome the high-computation requirement of dense method by exploiting pixels only with strong gradients (i.e., edges) [23]. Although they can retrieve a more complete description of the environment compared with sparse SLAM, however, in principle, being able to use all the information in the images should provide better estimates, as opposed to discarding a large number of pixels, even if they are less-informative. So, it is meaningful to find a simple and lightweight method to make the semi-dense reconstruction denser with lower computational cost.

### 2.2 Multispectral Photometric Stereo

Photometric Stereo is a popular method for 3D reconstruction from images due to its high level of details handling. It recovers 3D shapes from fixed viewpoint with more than three different artificial illuminations [24, 3, 25]. However，traditional photometric stereo needs to have a set of light sources, and capture an image separately under each light source and it needs to know the lighting direction of each light source. Drew et al. [26] and Kontsevich et al. [27] initially proposed a multi-spectral photometric stereo technique, which can obtain a detailed geometry structure from a single image. In essence, multi-spectral photometric stereo is photometric stereo with colored light. Unlike photometric stereo which photographs objects under varying white lights and processes gray-scale images, the multi-spectral photometric stereo captures an RGB image, which stores pixels as one byte each for red, green, and blue values, under three colored light sources at one time.

Using depth as prior information can significantly improve the accuracy of 3D reconstruction. Commercial depth sensors such as Kinect and Real Scene can acquire three-dimensional information of objects in real time without the need to know objects or lighting in advance.

For example, Zhang et al. [29] and Yu et al. [30] introduced several sensor fusion schemes that combine active stereo with photometric stereoscopy. [31] uses depth data comes from Kinect for multispectral photometric stereo. However, these methods are highly dependent on depth sensor and gratly limits the algorithm's scope of application. Be inspired that SLAM algorithms estimates depth maps when it performs localization and mapping, we can import depth map from SLAM into multi-spectral photometric algorithm for 3D shape recovery. Considering that the experimental setup of [9] exactly meets our experimental requirements, that is, they need to reconstruct the objects' three-dimensional structure filmed by a dynamic camera, we can take advantage of the mothods they use. In contrast, we use depth map from LSD-SLAM as priors in multispectral photometric stereo algorithm for dense 3D reconstruction.

### 2.3 Hybrid algorithms for 3D reconstruction

There are some hybrid approaches for more accurate dense 3D reconstruction. Some methods combine positions and surface normals by formulating the reconstruction problem as the optimization of the constraints that different measurements provide, for example, the study that Chen et al. [32] performed. However, this type of optimization methods may be too expensive. Other approaches first obtain a surface by integrating normals, and then combine the obtained mesh with measured positions. Nevertheless, these approaches may introduce bias and are not robust [33]. [16] employs an efficient linear optimization method based on the sparse, global depth information produced by using encoded structured light to correct the bias that photometric stereo yields. However, none of these methods are video-based, and they have great limitations in practical applications. [13] solves for the transformation between sparse 3D surface estimation from SFM and dense 3D localization from PS to correct alignment of the two reconstructions up to an overall scale and finally get a more accurate dense reconstruction result. The method comprehensivly consider the mutiview geogmetry and photometric prospective. Although the paper mentions that their experiments were performed on video sequences, their images are taken every time the object is fixed to a specific angle, arranged in an unordered manner which results non-real-time running, rather than our video that could observe the camera motion frame by frame. Similar with their method, we also use knowledge of muti-view geogmetry and photometric prospective. The difference is that we perform our experiment in real time in still scenes with a moving camera.

## 3 METHOD

In this section, we illustrate the proposed framework based on mutispetral data for dense 3D reconstruction, (shown in Fig.2) where dense surface normals using multispectral photometric stereo are fused together with semi-dense 3D model obtained from direct monocular SLAM. The three main components of the framework are then described in Sec. 3.1 (SLAM), Sec. 3.2 (dense surface normal recovery), Sec. 3.3(fusion procedure).

### 3.1 SLAM

We employ a key-frame based SLAM paradigm [11,17], in particular we use as baseline the direct semi-dense approach in [17]. [17] uses the direct method to operate in the high-gradient region of the keyframe image to reconstruct a semi-dense point cloud. The global reconstruction result collects information of all keyframes.

During tracking and mapping in [17], we can obtain depth maps of keyframes and camera pose. Semi-dense depth maps are obtained by filtering over many pixel-wise stereo comparisons. Once a new frame is chosen to become a keyframe, its depth map is initialized by projecting points from the previous keyframe into it, followed by one iteration of spatial regularization and outlier removal. Tracked frames that do not become a keyframe are used to refine the current keyframe until the next keyframe is selected. When the new keyframe is selected, the depth map of the current keyframe has been optimized to an optimal value.

For depth map extraction, we found that in [17], each key-frame is associated with an inverse depth map. In this inverse depth map, each pixel corresponds to an inverse depth value. We take the inverse of the inverse depth value as the depth of each pixel. For a pixel with a negative inverse depth value which means the depth estimated is invalid, we set its depth value to 0; for a pixel with no corresponding inverse depth value, we also set its depth value to 0. In this way, we can obtain initial depth maps of keyframes from SLAM for further dense surface normal recovery via multispectral photometric stereo.

Camera pose estimation is carried out at each input frame. During tracking, by estimating the transformation between the current frame and its nearest key-frame, camera pose denotes camera's position and angle transformation in 3D, and can be defined as

$$G = \begin{pmatrix} R & t \\ 0 & 1 \end{pmatrix} \text{with } R \in SO(3) \text{ and } t \in \mathbb{R}^3 \ . \quad (1)$$

Where $R$ denotes the rotation metrix and $t$ denotes the translation vector. In the process of mapping, a scaling factor s is introduced to adjust the camera's rotation matrix for image alignment and the camera pose subject to global refinement based on pose graph optimization is defined as

$$S = \begin{pmatrix} sR & t \\ 0 & 1 \end{pmatrix} \text{with } R \in SO(3), t \in \mathbb{R}^3 \text{ and } s \in \mathbb{R}^+ \quad (2)$$

Note that in [17], as for camera pose corresponding to the global 3D reconstruction, a minimal representation is given by elements of the associated Lie-algebra $\xi \in \text{sim}(3)$, which have 7 degree of freedom, that is $\xi \in R^7$. Here, the rotation matrix is represented as the quaternion $(q_w, q_x, q_y, q_z)$. In our work, we use the camera pose represented as a 3 × 4 transform matrix which is formed by a 3×3 rotation matrix and 3×1 translation vector. So, we get the 3× 3 rotaion from the quaternion by:

$$R = \begin{bmatrix} q_w^2 + q_x^2 - q_y^2 - q_z^2 & 2(q_x q_y - q_w q_z) & 2(q_x q_z + q_w q_y) \\ 2(q_x q_y + q_w q_z) & q_w^2 - q_x^2 + q_y^2 - q_z^2 & 2(q_y q_z - q_w q_x) \\ 2(q_x q_z - q_w q_y) & 2(q_y q_z + q_w q_x) & q_w^2 - q_x^2 - q_y^2 + q_z^2 \end{bmatrix} \quad (3)$$

Then the transform matrix representing camera pose we obtained from SLAM is used for subsequent view conversion.

### 3.3 dense surface normal recovery

The multi-spectral photometric stereo algorithm has the advantages that a three-dimensional model requires only one color image to reconstruct 3D model, and thus can be used for video reconstruction problems. In SLAM algorithm, eachtime a keyframe is published by SLAM, we calculate its dense surface normal using mutisectral photometric stereo by incoprating the depth maps from SLAM as priors.

The principle of multi-spectral photometric stereo is shown in Equation (3):

$$c_i(x, y) = \sum_i l_j^T n(x, y) \int E_j(\lambda) R(x, y, \lambda) S_i(\lambda) d\lambda \quad (4)$$

Where, $lj$ is the j-th illumination direction vector, $n(x, y)$ is the normal vector of a certain point of the target, $E_j(\lambda)$ is the illumination intensity, $R(x, y, \lambda)$ is a parameter related with the albedo and chromaticity of a certain point of the target, and $S_i(\lambda)$ is the color response of the camera photosensitive element.

Assume $R(x, y, \lambda)$ as the product of $\rho(x, y)$ and $\alpha(\lambda)$ in Equation (3) which represents the albedo and the chromaticity respectively, then put all items which are related with $\lambda$ as a whole, and we can get a parameter matrix V, as shown in Equation (5):

$$V_{ij} = \int E_j(\lambda) \alpha(\lambda) S_i(\lambda) d\lambda \quad (5)$$

So, we can rewrite Equation (4) as Equation (6):

$$C = VL\rho n \quad (6)$$

According to Equation (6), if we assume that there is a matrix M,

$$M = VL\rho \quad (7)$$

M can be estimated from multispectral keyframes. Disadvantage of this method is that the surface color, albedo, light source information, and camera sensor information of different objects are confusing. So, we use depth map extracted from SLAM (described in section **3.1**) as priors to improve accuracy of reconstruction.

We can get gradient g from the optimized depth maps, and M can be definded as

$$M = Cg \quad (8)$$

So, using M optimized by g, the surface normal of the object can be computed by

$$n = M^{-1}C \quad (9)$$

During dense surface normal recovery, considering the objects with multi-chromaticity, we use method proposed in [9], where the sections with the same surface chromaticity and albedo are segmented using SILC algorithm and the cooresponding normal are calculated respectively for further normal recovery.

### 3.4 Fusion Procedure

We combine semi-dense point cloud from SLAM with dense surface normal from multispectral photometric stereo in our fusion procedure. The main process of the fusion procedure in this paper composes three steps: point cloud view conversion, fusion and joint optimization, point cloud registration.

**Point cloud view conversion.** The 3D model of SLAM contains information of all key frames of each view, and the multi-spectral photometric stereo method obtains the pixel-level normal information at a specific view. If they are to be combined, they must first ensure that they are in the same view. Therefore, we need to transform point cloud view. Using camera pose obtained from section **3.1**, we multiply the camera transition matrix with the point cloud model and derive the point cloud of the current key frame view.

**Fusion and joint optimization.** We insert the dense surface normal into semi-dense point cloud, so that we can obtain a preliminary fusion result. The result includes surface normal with high-frquency information and point cloud coordinates with low-frequency information. However, the high-frequency information and the low-frequency information are independent from each other, so a joint optimization of the point cloud and the surface normal is needed. The joint optimization of coordinates and normals is a mathematical optimization problem.

Rushmeier et al. proposed the coordinate normal optimization algorithm based on the fusion of multi-view stereo and photometric stereo [34], but this method will produce overlapping normals on the local surface. In [35], the author first uses the coordinate information to optimize the normal, then uses the normal information to optimize the coordinates, and obtains the final model by minimizing the global energy function. We adopt the method in [35], in which the optimization of normal and position has its weight coefficient that can be manually adjusted according to the effect of the previous semi-dense reconstruction of SLAM and the dense surface normal recovery of multispectral photometric stereo method.

**Point cloud registration.** Through above method, we get the optimized fusion result, dense point cloud. When the normal direction of a single keyframe reconstruction is not ideal, for example, there is no normal information in some areas of the normal graph, which will lead to a larger error in the fusion of surface normal with the point cloud. We use the ICP algorithm to register the point clouds that have been merged with different key frames, and to complete the areas where there is no normal information in the point cloud.

## 4 EXPERIMENTS

In this section, we will discuss our experiments and give experiment results in 4 parts: Multispectral Setup, Results of Multispectral Photometric Stereo, Results of Fusion, and Point Cloud Registration.

### 4.1 multispectral setup

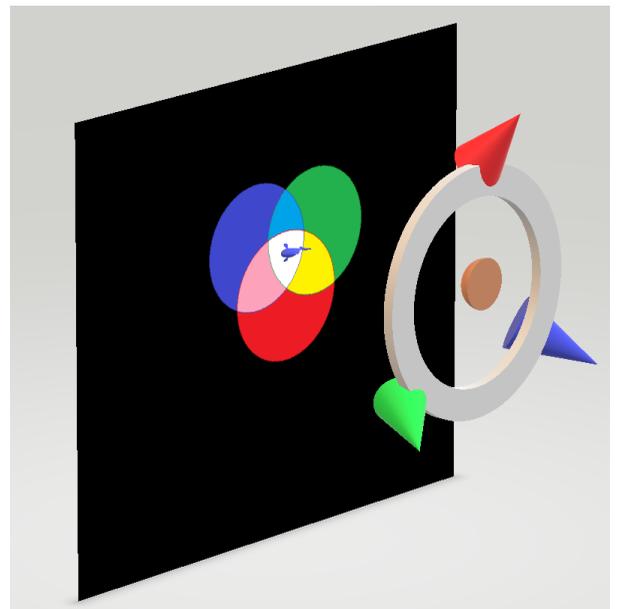

Fig. 3. A schematic representation of our multispectral setup. The three cones represent red, green and blue light sources respectively, and the brown semicircle represents the camera.

The proposed framework performs on multispectral data.

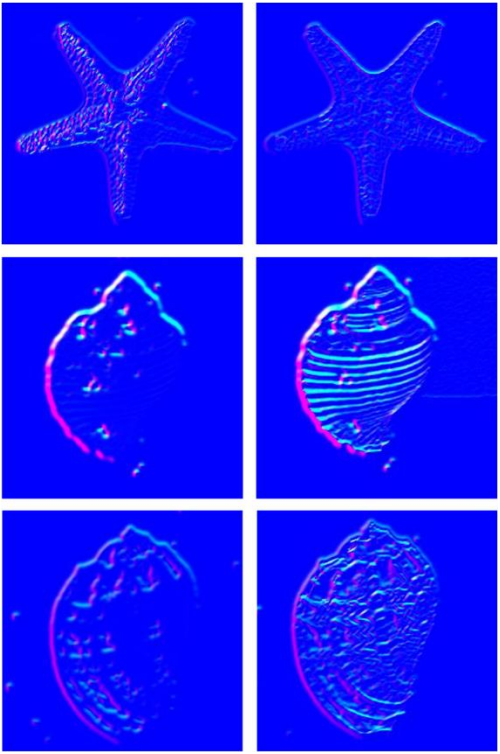

Fig. 4. Dense surface normal before and after optimization. Left: surface normal before optimization. Right: surface normal after optimization.

will reflect each of those colors simultaneously without any mixing of the frequencies. The quantities of red, green and blue light reflected are a linar function of the surface normal direction. A color camera can measure these quantities, from which an estimate of the surface normal direction can be obtained. During the process of acquiring, the relative position of the camera and the lights is fixed, camera and three light sources, as a whole, move around the object, and the object is photographed. In order to avoid the interference of ambient light, our experiments were conducted in a dark environment.

In our experiment, we resize all the images to 512×512. In terms of SLAM, which needs a high frame-rate, our frame-rate is 30fps.

### 4.2 Results of Multispectral Photometric Stereo

We calculate dense surface normal using multispectral photometric stereo which needs denpth map as priors. However, the initial depth maps extracted from SLAM are not accurate and dense, so we perform optimization on them. Considering that the inverse depth map has been scaled to have a mean inverse depth of one in SLAM, it is necessary to multiply the depth value with the scaleing factor to obtain its original depth value. Note that the depth map and variance are only defined for a subset of pixels containing all image regions in the vicinity of sufficiently large intensity gradient, hence semi-dense. So,

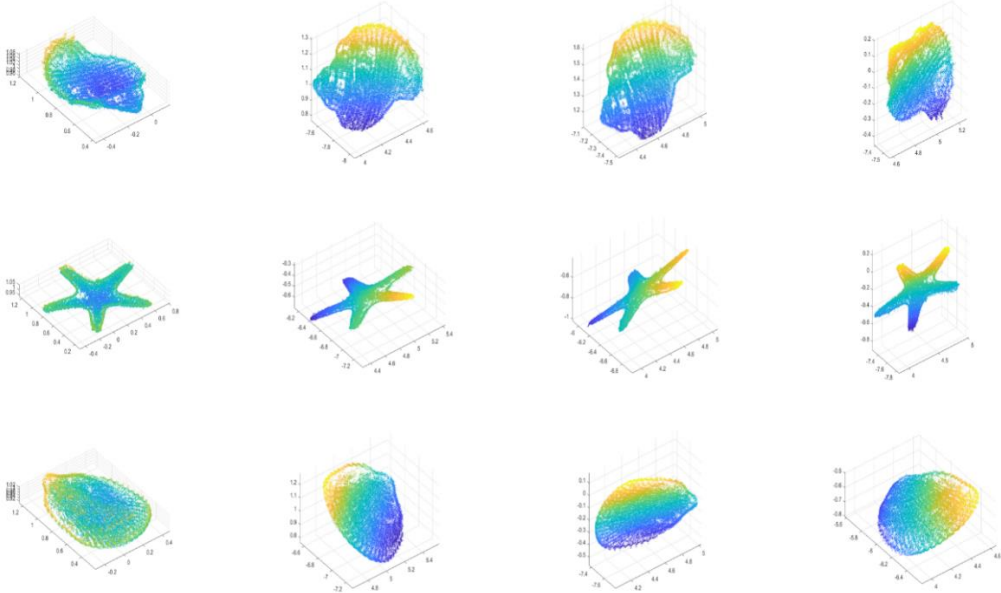

Fig. 5. View conversion of point cloud. The left column is the point cloud of the original view, and the right three columns are the point cloud transformed using different camera pose matrix.

For acquiring multispectral video, we use a practical set-up that consists of an industrial video camera and three colored light sources (see Fig. 3). The key observation is that in an environment where red, green, and blue light is emitted from different directions, a Lambertian surface we use bilinear interpolation to fill hole of the depth maps to make it denser. Afterwards, we will derive more accurate and dense depth maps.

Then we use the opyimized depth maps to perform dense normal recovery.

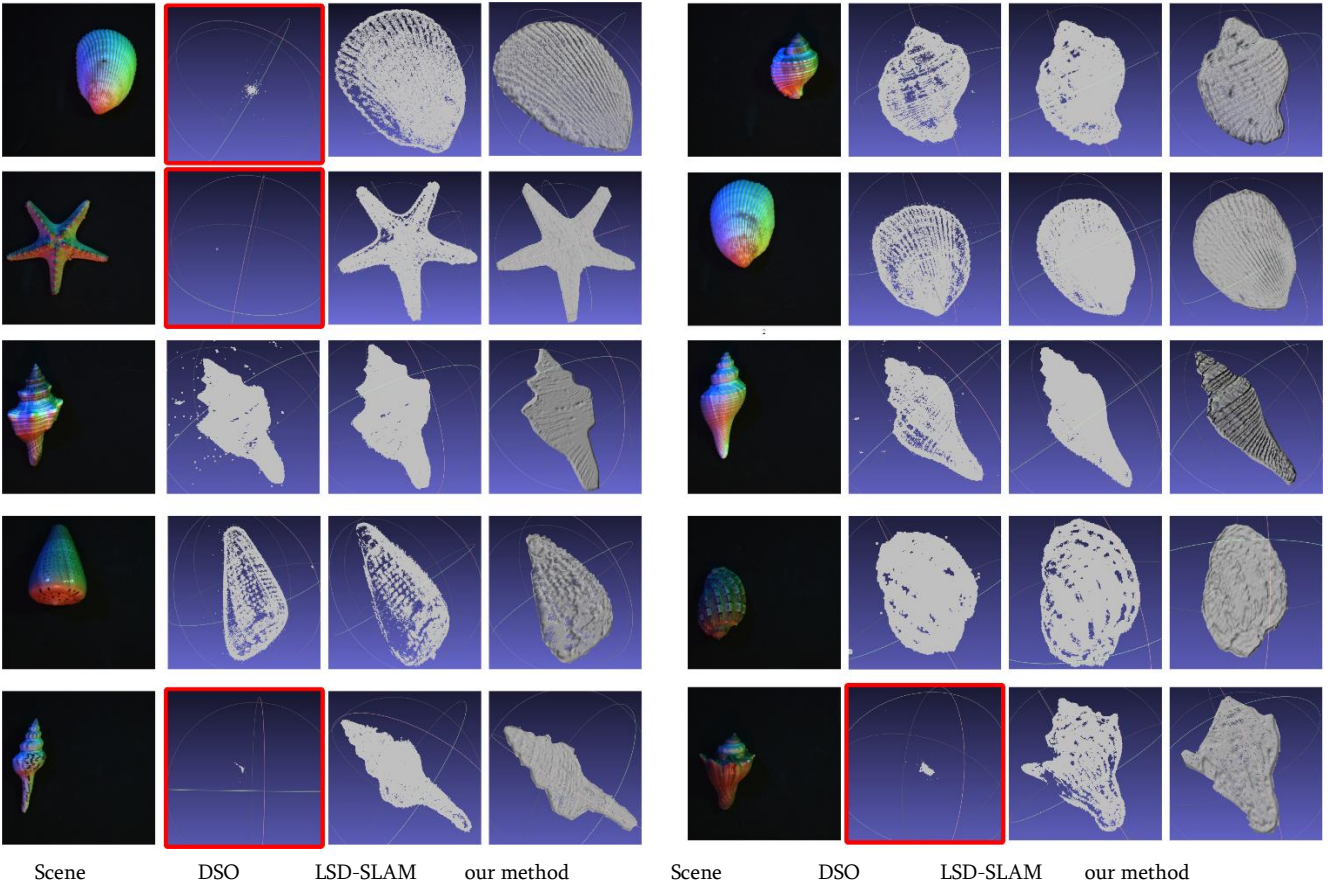

Fig. 6. Results of fusion. In the above eight groups of images, the left is the scene we shot. The three columns on the right are reconstruction results of DSO, LSD-SLAM, and our method respectively. Note that in the first and second rows, the scene marked with a red rectangle is a failed reconstruction. The reason is that the information collected by dso is too sparse to get the reconstruction result.

We show the dense surface normal before and after optimization in Fig.4. As can be seen from the figure, our optimization improves the reconstruction that texture recovery is more successful.

### 4.3 Results of fusion

We combine the semi-dense point cloud build from SLAM and dense surface normal calculated from multi-spectral photometric stereo to make the semi-dense reconstruction denser and more accurate.

In our experiment, we only use the position information in the point cloud to merge with the surface normal, because the normal direction obtained using the multi-spectral photometric stereo is per-pixel.

Because the spaces of obtained semi-dense point cloud and the dense normal direction do not have a unified unit of measure, we need to perform area matching by using view conversion and scaling. First, we convert the global point cloud in order to ensure that its view is consistent with the current keyframe's. We use 3×4 camera pose described in section **3.1** to convert the view so that the X0Y plane of the coordinate system of the point cloud is parallel to the plane of the normal to facilitate the subsequent mapping of the two types of information.

Then we multiply the camera's transform matrix with the global point cloud of SLAM to obtain the point cloud whose view is consistent with the current keyframe's. Fig.5 shows the result of view convertion.

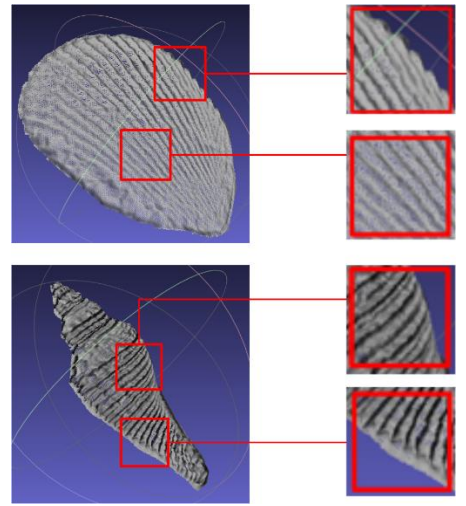

Fig. 7. Some details of fusion results.

For scaling, we scale the normal map with a linear map method to align it with the corresponding area of the point cloud at the same view.

Then, we map the normal information to the corresponding position of the point cloud, so that the points that have not been sampled in the point cloud have the initial geometric information. We store the normal information and position information of the point cloud at the same time to get the initial fusion result.

In order to unify the normals with high-frequency information and the point clouds with low-frequency information into a whole, rather than independent of each other in initial fusion result, we use the method in [35] to optimize the normal and point clouds. [35] is motivated by an analysis of the common error characteristics of measured normals and positions. The method proceeds in two stages: first, it corrects for low-frequency bias in the measured normal field with the help of measured surface positions. Then, it optimizes for the final surface positions using linear constraints and an efficient sparse solver. In addition, only the most reliable frequency components of each source are considered, resulting in a reconstruction that both preserves high-frequency details and avoids low-frequency bias. During optimization, normal and position each has its own weight coefficient, and can be manually adjusted according to the reconstruction quality using SLAM and multispectral photometric stereo. Through a number of experiments, we found that setting the weight coefficient to 1:3 can obtain better experimental results. Fig.6 shows the result of fusion. We compare several CPU-based SLAM algorithms (DSO and LSD-SLAM) with our method. We can see that with simi-lar computational costs, our method obtained denser point clouds. Table 1 shows the comparison of running time under several videos of different durations. Running times of LSD-SLAM, DSO and our method will be slightly different each time on the same video. In order to avoid the contrast unfairness caused by this small randomness, we use the three method to run 10 times for each video, record their running time, and then take the average of 10 running times to get the running time in the **Table1**. We can see that our framework runs only slightly slower than DSO and LSD-SLAM. In Figure 7, we show some details of the fusion results obtained via our method. From the figure, we can see that our results have a fine texture, which indicates that our method can capture more high-frequency information. In Figure 8, we also compare the three-dimensional structure of the point cloud obtained from SLAM and our results. We can see that by fusing the normal information and joint optimization, our method has improved the position information of the point cloud obtained from SLAM, access to a realistic three-dimensional structure

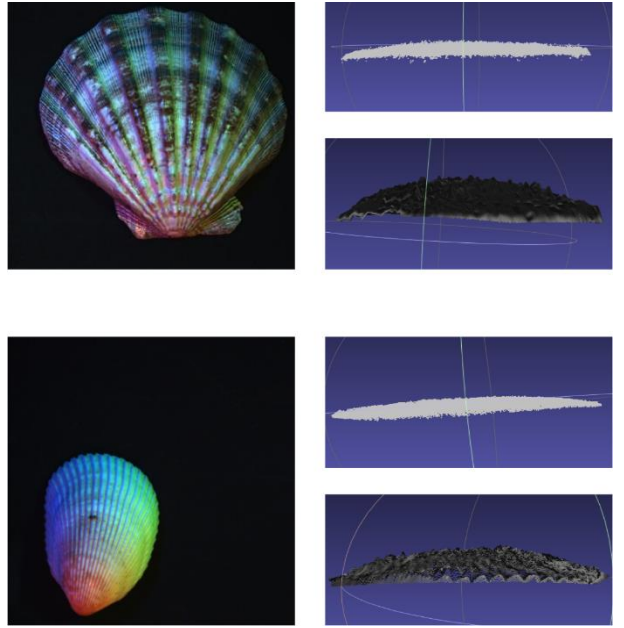

Fig. 8. Position information optimized.

Table 1. Running time of several video durations.

| Video durations(frame) | DSO | LSD-SLAM | Our method |
| --- | --- | --- | --- |
| 1379 | 130s | 146s | 160s |
| 1471 | 136s | 154s | 168s |
| 1524 | 151s | 161s | 170s |
| 1680 | 165s | 173s | 186s |

### 4.4 Point Clouds registration

In most cases, we can obtain better fusion results through the above steps. However, sometimes the normal information in the normal direction reconstructed by the multi-spectral photometric stereo algorithm will be partially missing, resulting in less than ideal fusion results. We use ICP algorithms for further optimization.

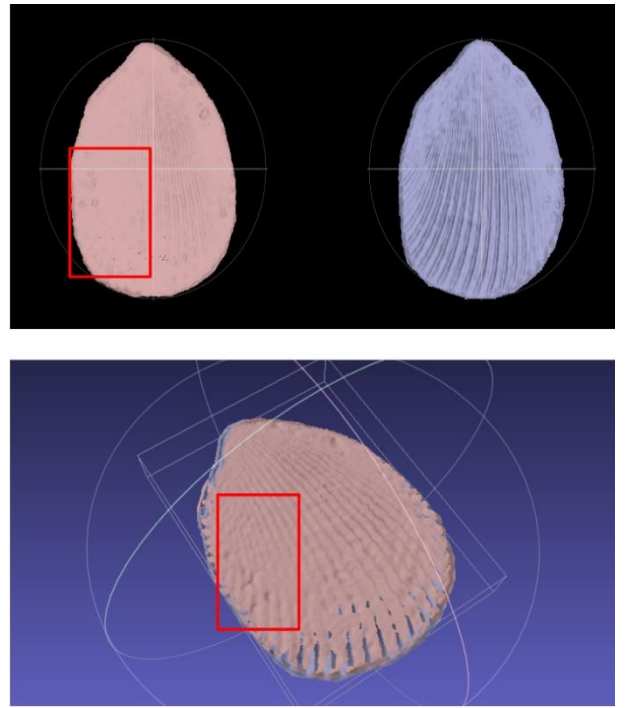

Fig. 9. Results of ICP algorithm.

The main steps of ICP algorithm include searching for common key points, generating feature descriptors, feature matching, and estimating the transformation ma-trix of the point cloud. The ICP algorithm finds the overlap between the various point cloud models and uses these overlapping as key points in the subsequent estimation transformation matrix. The feature points are generated using the key points found, and the matching between the point clouds is performed according to the feature descriptors.

Through the ICP algorithm, we can integrate the information of overlapping parts belonging to multiple normal graphs to fill the missing information and obtain more accurate results. The results of algorithm are shown in Fig. 9. The upper left image shows the reconstruction with partial tex-ture noise caused by the lack of normal information. The bot-tom is the result of the registration of the two point clouds above using the ICP algorithm. Note that the point cloud at the bottom has two colors, which represents the information of the registered point cloud from two different point clouds. From the figure, we can see that the ICP algorithm can improve the reconstructin of texture in the noise area.

## 5 CONCLUSION

We have shown how we combine the two approaches，monocular semi-dense SLAM and multispectral photometric stereo to rebuild a dense 3D reconstruction in real time where the multispectral data is shared by SLAM and multispectral photometric stereo algorithms. To the best of our knowledge, this is the first demonstration of joint multispectral photometric stereo and SLAM with a monocular camera for dense 3D reconstruction. Note that our framework runs on two parallel systems and can reconstruct the three-dimensional structure of the scene in real time on CPU. Experiments demonstrate that our framework is capable of attaining real-time dense 3D reconstruction.

As a next step, we plan to generate higher quality 3D models based on the optimized depth maps such as [36], where the depth prediction via a deep neural network is integrated with SLAM.

10[23] Kim, Jae Hak, C. Cadena, and I. Reid. "Direct semi-dense SLAM for rolling shutter cameras." IEEE International Conference on Robotics and Automation IEEE, 2016:1308-1315.

[24] Woodham, R. J. "Photometric Method for Determining Surface Orientation from Multiple Images." Optical Engineering 19.1(1980):1-22.

[25] Dong, Junyu, et al. "Improving photometric stereo with laser sectioning." IEEE, International Conference on Computer Vision Workshops IEEE, 2009:1748-1754.

[26] Drew, M. S., and L. L. Kontsevich. "Closed-form attitude determination under spectrally varying illumination." Computer Vision and Pattern Recognition, 1994. Proceedings CVPR '94. 1994 IEEE Computer Society Conference on IEEE, 1994:985-990.

[27] Petrov, A. P., I. S. Vergelskaya, and L. L. Kontsevich. "Reconstruction of shape from shading in color images." Journal of the Optical Society of America A 12.11(1994):1047-1052.

[28] Zhang, Q.; Ye, M.; Yang, R.; Matsushita, Y.; Wilburn, B.; Yu, H. Edge-preserving photometric stereo via

[29] depth fusion. In Proceedings of the IEEE Conference on Computer Vision and Pattern Recognition (CVPR), Providence, RI, USA, 16–21 June 2012; pp. 2472–2479.

[30] Yu, Lap Fai, et al. "Shading-Based Shape Refinement of RGB-D Images." Computer Vision and Pattern Recognition IEEE, 2013:1415-1422.

[31] Seitz, Steven M., R. Szeliski, and N. Snavely. "Photo Tourism: Exploring Photo Collections in 3D." Acm Transactions on Graphics 25.3(2006): págs. 835-846.

[32] Chen, Chia Yen, R. Klette, and C. F. Chen. Shape from Photometric Stereo and Contours. Computer Analysis of Images and Patterns. Springer Berlin Heidelberg, 2003:377-384.

[33] Mostafa, Mostafa G. H., S. M. Yamany, and A. A. Farag. "Integrating Shape from Shading and Range Data Using Neural Networks." Computer Vision and Pattern Recognition, 1999. IEEE Computer Society Conference on IEEE, 1999:2015-2020.

[34] Rushmeier, H., and F. Bernardini. "Computing consistent normals and colors from photometric data." International Conference on 3-D Digital Imaging and Modeling, 1999. Proceedings 2002:99-108.

[35] Nehab, Diego, et al. "Efficiently combining positions and normals for precise 3D geometry." Acm Transactions on Graphics 24.3(2005):536-543.

[36] Tateno, Keisuke, et al. "CNN-SLAM: Real-Time Dense Monocular SLAM with Learned Depth Prediction." (2017):6565-6574.